\documentclass[runningheads]{llncs}
\usepackage[T1]{fontenc}
\usepackage{graphicx}
\usepackage{amsmath}
\usepackage{amssymb}
\usepackage{amsfonts}
\usepackage{algorithmic}
\usepackage{algorithm}
\usepackage{appendix}
\usepackage{comment}

\newcommand\blfootnote[1]{%
  \begingroup
  \renewcommand\thefootnote{}\footnote{#1}%
  \addtocounter{footnote}{-1}%
  \endgroup
}

\begin{document}
\title{OnDeFog: Online Decision Transformer under Frame Dropping}
%
%
\author{Daiki Yotsufuji \and
Kenta Nishihara \and
Shoma Shimizu \and
Kento Uchida \and
Shinichi Shirakawa
}
\authorrunning{Y. Daiki et al.}
%
\institute{Yokohama National University,
Kanagawa, Japan
}
%
\maketitle              
\begin{abstract}
In challenging real-world reinforcement learning applications, communication delays or sensor failures often cause frame dropping, in which the agent cannot receive the dropped states and associated rewards. To address the performance degradation caused by frame dropping, the Decision Transformer under Random Frame Dropping (DeFog) was developed by incorporating additional mechanisms into the decision transformer to tackle frame dropping. Although DeFog can mitigate performance degradation in frame-dropping environments, since DeFog is an offline learning method, it struggles to effectively generalize to novel states not adequately represented in the training dataset. In this study, we propose OnDeFog, which integrates the mechanisms in DeFog with the online decision transformer (ODT), an online reinforcement learning method that learns policies through direct environmental interaction. Comprehensive experimental evaluation demonstrates that our proposed OnDeFog achieves superior performance compared to ODT in environments characterized by high dropping frame rate and outperforms DeFog on datasets containing a large amount of low-reward data.
\blfootnote{Our experimental code is available at \url{https://github.com/shiralab/OnDeFog}.}
\end{abstract}
\section{Introduction}
Reinforcement learning (RL) has been applied and succeeded in various fields, such as playing games, controlling drones, and driving autonomous vehicles~\cite{shao2019survey,siedler2022dynamic}. One of the challenges in deploying RL in real-world environments is incomplete observations (e.g., missing states or rewards) during agent-environment interactions. Communication delays or sensor failures, often referred to as frame dropping, cause incomplete observations, leading to the performance degradation of RL~\cite{dulac2021challenges}. To address this issue, methods for environments prone to frame dropping have been investigated~\cite{bouteiller2021reinforcement,chapman1991input,hudecision,nath2021revisiting}.

Particularly, Hu et al. \cite{hudecision} proposed the Decision Transformer under Random Frame Dropping (DeFog) to enable robust agent control in environments prone to frame dropping based on the Decision Transformer~\cite{chen2021decision}, a method of offline RL. DeFog explicitly learns to handle frame dropping, achieving significantly higher robustness than conventional methods. A limitation of DeFog is that the learned policy may fail to select appropriate actions in situations not present in the dataset because DeFog is an offline learning method relying on pre-collected datasets.
Consequently, DeFog requires preparing high-quality and extensive datasets in practice, resulting in high dataset collection costs.

In contrast, online RL learns policies through direct interaction with the environment, enabling learning even when pre-collected datasets are insufficient. Online Decision Transformer (ODT)~\cite{zheng2022online} extends the Decision Transformer to online RL and successfully improves performance on tasks where the conventional Decision Transformer works poorly through online fine-tuning. However, ODT itself is not designed to handle frame dropping, raising concerns about performance degradation when frame dropping occurs.

Addressing these challenges, this study proposes Online DeFog (OnDeFog), which integrates DeFog's mechanism for handling frame dropping into online RL. OnDeFog aims to enhance robustness against frame dropping in real-world environments by incorporating techniques used in DeFog into the online learning framework of ODT. Experimental results show that OnDeFog has superior performance to ODT in high-drop-rate environments and outperforms DeFog on low-reward datasets.

\section{Related Work}

\subsection{Offline Reinforcement Learning}
Offline RL trains an agent using a pre-collected dataset without direct environmental interaction~\cite{levine2020offline}, in contrast to online RL, which learns through direct interaction. A primary advantage of offline RL is safety, as it avoids executing potentially hazardous actions during real-world training. However, a major challenge is its susceptibility to biases within the dataset, which can prevent the policy from learning effective actions for states not well-represented in the training data~\cite{fu2020d4rl}.

\subsection{Robust RL Methods Against Frame Dropping and Delay}
Frame dropping and delay are perennial and critical challenges in communication-dependent teleoperation tasks. Simon et al.~\cite{240649} proposed a Soft Actor-Critic (SAC)~\cite{haarnoja2018soft}-based method that leverages delayed observation information to evaluate past actions accurately. Nath et al.~\cite{4792652} proposed a Deep Q-Network (DQN)~\cite{DBLP:journals/corr/MnihKSGAWR13}-based RL framework to handle observation delays. However, these methods presuppose a fixed maximum duration for frame dropping and delay. Consequently, their performance may be suboptimal when frame dropping and delay extend beyond these predefined maximums.
The method most related to this study is DeFog~\cite{hudecision}, which is a robust offline RL method against frame dropping.
Because DeFog probabilistically defines frame dropping, such predefined maximums do not exist. 
Its detailed explanation is found in Section~\ref{sec:defog}.

\section{Preliminaries}
The interaction between an agent and the environment in RL is commonly formalized as a Markov Decision Process (MDP), defined by a 5-tuple $\mathcal{M}=\left\langle\mathcal{S}, \mathcal{A}, P_0, P, r\right\rangle$, where $\mathcal{S}$ is a state space, $\mathcal{A}$ is an action space, $P_0$ is an initial state distribution, $P(s'|s,a)$ is a state transition probability, and $r(s,a,s')$ is a reward function.
Here, we introduce the cumulative reward at step $t$ in RL, denoted as $\sum^T_{t'=t} r\left(s_{t'}, a_{t'}, s_{t'+1}\right)$, where $T$ represents the maximum episode length. RL aims to train a policy $\pi$ that maximizes the expected cumulative reward over multiple interactions, often referred to as the expected return, $\mathbb{E}_{\pi}\left[\sum_{t'=0}^{T}  r\left(s_{t'}, a_{t'}, s_{t'+1}\right) \mid s_0 \sim P_0 \right]$.

In the following sections, we describe methods relevant to this study: Decision Transformer, DeFog, and Online Decision Transformer.

\subsection{Decision Transformer} \label{sec:DT}
Decision Transformer (DT)~\cite{chen2021decision} is an offline RL technique that adapts the Transformer architecture~\cite{vaswani2017attention}, originally developed for sequential data processing, to RL tasks.
Compared to conventional RL methods, DT offers a significant advantage in terms of training stability.
While conventional RL methods aim to maximize cumulative reward, DT performs supervised learning by conditioning on a target return and predicting actions that lead to it.
Specifically, during training, DT learns to determine actions conditioned on the returns-to-go, which is defined as the sum of future rewards \(\sum^T_{t'=t}r_{t'}\).
During inference, DT cannot leverage future rewards from a given sequence of past time steps. Therefore, DT replaces the returns-to-go with the difference $g_t = g_{\mathrm{target}}-\sum^t_{t'=0}r_{t'}$ between a pre-specified target return $g_{\mathrm{target}}$ and the cumulative rewards observed up to that point, where $g_{\mathrm{target}}$ is a hyperparameter in DT.
This sequence of returns-to-go, along with the corresponding sequences of states and actions, is then input to the model.
This approach mitigates conventional methods' issues, such as accumulating value function estimation errors (also known as the bootstrapping problem) and premature convergence to local optima. Furthermore, adopting the Transformer architecture effectively captures long-range dependencies within state, action, and reward sequences, which facilitates the learning of sophisticated behavioral strategies that consider long-term context.

\subsection{DeFog}
\label{sec:defog}
DeFog~\cite{hudecision} is an offline RL method that exhibits high robustness against frame dropping.
This method mitigates performance degradation in frame-dropping environments through three key innovations: train-time frame dropping, drop span embedding, and freeze-trunk fine-tuning.
In train-time frame-dropping, the method simulates frame-dropping environments by intentionally dropping inputs (states and returns) at specific time steps within the dataset during training.
The previous state and return are carried over at these dropped time steps, enabling the model to acquire robustness to frame loss.
Drop span embedding involves calculating the number of consecutively dropped frames at each time step. This information is then encoded into a vector by a drop span encoder and added to the state and return tokens, enabling the model to explicitly learn to handle missing patterns.
After the initial model training, the freeze-trunk fine-tuning fixes most of the parameters, and only the drop span encoder and the neural network responsible for action inference continue to be trained. This approach achieves improved performance in environments with high drop rates.
Note that DeFog assumes to be able to access an offline dataset without frame dropping and realizes obtaining a robust policy for frame dropping in testing environments.

\subsection{Online Decision Transformer}
Online Decision Transformer (ODT) \cite{zheng2022online} is an extension of DT designed for online RL. It achieves efficient learning by first performing offline pre-training like DT, followed by online fine-tuning.
Unlike the original DT, which is an offline RL method and thus inherently unable to explore situations not present in the dataset, ODT introduces a stochastic policy. This enables it to explore unknown states through interaction with the environment. Furthermore, ODT employs a regularization that considers the entropy of the entire sequence of input states and rewards, rather than the policy entropy at all time steps.
This mechanism facilitates exploration at time steps where actions do not significantly harm the performance while preserving critical action fidelity.
During online fine-tuning, experiences collected through interaction with the environment are stored in a replay buffer as whole trajectories, where the trajectory indicates the sequence of state, action, and return-to-go $(s_t, a_t, g_t)$.
Additionally, ODT replaces the target return specified during the action decision with the actual return obtained. This corrects the discrepancies between the specified target return and the observed return during online exploration, enabling more accurate training.

\begin{figure}[tb]
\begin{center}
    \includegraphics[width=0.75\columnwidth]{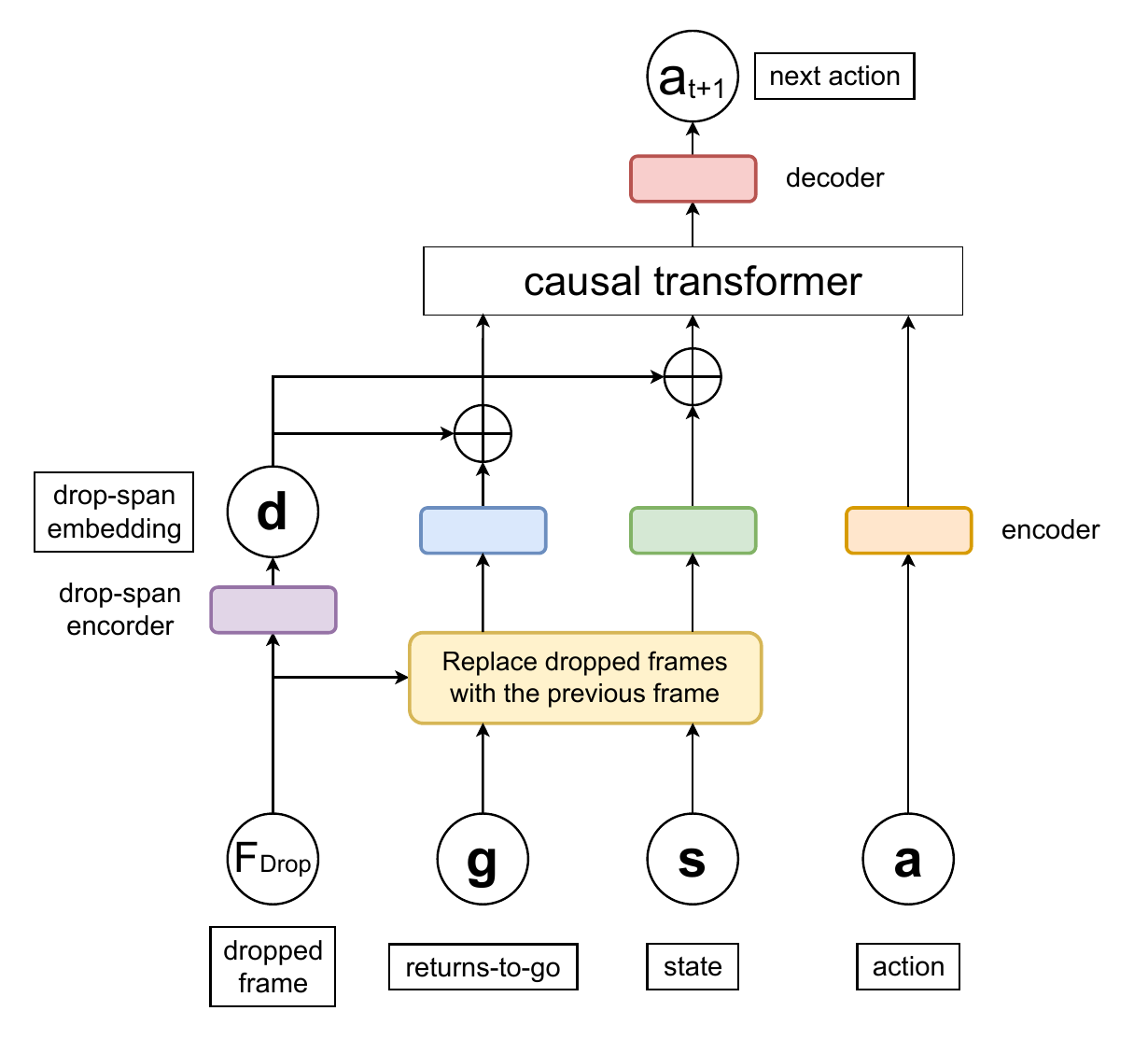}
    \caption{Overview of OnDeFog model}
    \label{model}
\end{center}
\end{figure}

\section{Proposed Method: Online DeFog}
In this study, we propose Online DeFog (OnDeFog), an online RL method designed to exhibit high robustness against frame dropping. OnDeFog integrates the DeFog technique into ODT, enabling stable training in frame-dropping environments, which were challenging for conventional ODT. Furthermore, OnDeFog mitigates DeFog's dependency on offline datasets by building upon ODT.
To illustrate how these advantages are realized, Figure~\ref{model} provides an overview of the model in the proposed OnDeFog. The inputs of OnDeFog are the set of dropped frames $F_{\mathrm{Drop}}$, the sequence of returns-to-go $\mathbf{g}$, states $\mathbf{s}$, and actions $\mathbf{a}$. The output corresponds to the predicted next action $a_{t+1}$.
The bold symbols herein indicate sequences.
Algorithms~\ref{OnDeFog} and \ref{OnDeFogTr} describe the details of the proposed method. 

\begin{algorithm}[t]
\caption{OnDeFog}
\label{OnDeFog}
\begin{algorithmic}[1]
\REQUIRE Offline data $\mathcal{T}_{\mathrm{offline}}$, number of exploration rounds $R$, target return for trajectory generation $g_{\mathrm{online}}$, replay buffer size $N$, number of updates $\{ I_{\mathrm{offline}}, I_{\mathrm{online}} \}$, policy $\pi_\theta$, dropping rates $p_{\mathrm{train}}$
\STATE Set dropping rate $p \leftarrow p_{\mathrm{train}}$ and number of updates $I \leftarrow I_{\mathrm{offline}}$
\STATE Pre-train using $\mathcal{T}_{\mathrm{offline}}$ (Algorithm~\ref{OnDeFogTr})
\STATE Initialize $\mathcal{T}_{\mathrm{{replay}}}$ with $N$ trajectories from $\mathcal{T}_{\mathrm{offline}}$ that have high returns
\STATE Set number of updates $I \leftarrow I_{\mathrm{online}}$
\FOR{$\mathrm{round}=1,\dots,R$}
\STATE Generate a sequence of tuples $(a_t,s_t,r_t)$ in the real environment according to policy $\pi_\theta(\mathbf{a}\mid \mathbf{s};g_{\mathrm{online}})$
\STATE Construct the trajectory $\tau$ using $(a_t,s_t,g_t)$ with the returns-to-go $g_t=\sum^{|\tau|}_{j=t}r_j$ \label{line:rtg}
\STATE Select the oldest trajectory $\tau_{\mathrm{oldest}}$ stored in the replay buffer
\STATE $\mathcal{T}_{\mathrm{{replay}}}\leftarrow\{\mathcal{T}_{\mathrm{{replay}}} \setminus \{\tau_{\mathrm{oldest}}\}\}\cup\{\tau\}$
\STATE Fine-tune using $\mathcal{T}_{\mathrm{{replay}}}$ (Algorithm \ref{OnDeFogTr})
\ENDFOR
\end{algorithmic}
\end{algorithm}

\begin{algorithm}[t]
\caption{OnDeFog Policy Training}
\label{OnDeFogTr}
\begin{algorithmic}[1]
\REQUIRE Parameters $\theta$, replay buffer $\mathcal{T}$, number of updates $I$, dropping rate $p$, Batch size $B$, sub-trajectory length $K$
\STATE Calculate selection probability for each trajectory $P_{\mathrm{select}}(\tau)=|\tau|/\sum_{\tau'\in\mathcal{T}}|\tau'|$
\FOR{$t=1,\dots,I$}
\STATE Select a set of $B$ trajectories $\mathcal{T}_{\mathrm{select}}$ from $\mathcal{T}$ according to $P_{\mathrm{select}}(\tau)$
\FOR{$\tau \in \mathcal{T}_{\mathrm{select}}$}
\STATE $(\mathbf{a},\mathbf{s},\mathbf{g})\leftarrow$ training context with the length of $K$ uniformly sampled from $\tau$
\STATE Determine dropped frames $F_{\mathrm{Drop}}$ according to dropping rate $p$
\STATE For $k \in F_{\mathrm{Drop}}$, set $(a_{k},s_{k},g_{k})\leftarrow(a_{k},s_{k-1},g_{k-1})$
\STATE Compute drop span embedding $\mathbf{d}$ and embed it into the states $\mathbf{s}$ and the returns-to-go $\mathbf{g}$, respectively
\ENDFOR
\STATE $\theta\leftarrow$ Update parameters $\theta$ using the selected $B$ sub-trajectories $(\mathbf{a},\mathbf{s},\mathbf{g})$ toward the gradient direction of the negative log-likelihood loss
\ENDFOR
\end{algorithmic}
\end{algorithm}

OnDeFog's training process adopts a two-stage approach similar to ODT: first, a policy is trained using the pre-collected dataset via offline RL, and then the policy is further improved through online RL. For simplicity, we assume that no frame dropping occurs in the trajectories $\tau$ obtained during the online RL stage. This two-staged training enables efficient exploration while leveraging the pre-collected dataset. 
Under the assumption that optimal return for the task is positive, the target return $g_{\mathrm{online}}$ during online learning is set to twice the optimal return. 
This scaled target value encourages the agent to explore more aggressively and promotes diverse data collection. Furthermore, in line~\ref{line:rtg} of Algorithm~\ref{OnDeFog}, the actual measured return is used for learning instead of the difference between the target return and the sum of rewards. This mechanism is attributed to the hindsight return relabeling in ODT.

In this study, we incorporate two key techniques from DeFog into ODT to handle frame dropping during training: train-time frame dropping and drop span embedding. 
Freeze-trunk fine-tuning is not employed because it did not yield substantial results when implemented in OnDeFog (see Appendix~\ref{sec:suppl_result} for the corresponding results).

\subsection{Train-time Frame Dropping}
Train-time frame dropping, as shown in Algorithm~\ref{OnDeFog}, utilizes dropping rates $p_{\mathrm{train}}$ for pre-training.
In Algorithm~\ref{OnDeFogTr}, dropping is applied to each frame (except the first) with probability $p$ for a selected trajectory from the replay buffer. The state and reward of the preceding frame are carried over in the dropped frames.

\subsection{Drop Span Embedding}
Drop span embedding is utilized during the update of the policy parameters $\theta$, as shown in Algorithm~\ref{OnDeFogTr}. 
The number of consecutive occurrences of frame-dropping steps is defined as $\mathbf{D}$.
For instance, ${D}_t=0$ for frames where no dropping occurs, and ${D}_t=2$ if dropping occurs for two consecutive steps. Subsequently, $\mathbf{D}$ is converted into an embedding vector $\mathbf{d}$ using a drop span encoder. Then, the states and returns are transformed using each encoder, and the embedding vector of $\mathbf{d}$ is added to both the transformed state and return, respectively. This allows for explicit consideration of dropping during parameter updates and action inference.

\section{Experimental Evaluation}
\subsection{Tasks and Datasets}
The D4RL benchmark~\cite{fu2020d4rl}, a collection of offline datasets for various RL tasks, was utilized for the evaluation. Specifically, the Gym-MuJoCo datasets~\cite{1606.01540}, designed for continuous control tasks, were adopted from D4RL. While the Gym-MuJoCo benchmark encompasses a wide range of environments, three specific tasks were selected for this study: Hopper, Walker2d, and HalfCheetah. 

For comparative online RL experiments, we employed the official OpenAI Gym MuJoCo environments.
For offline RL, the D4RL~\cite{fu2020d4rl} medium and medium-replay datasets were utilized. 
The medium dataset comprises trajectories collected by a policy that achieves approximately one-third of the optimal policy's performance. The medium-replay dataset consists of replay data collected during the training process of an untrained policy aiming to reach a medium-level performance level. This medium-replay dataset is particularly challenging for training because it includes many low-reward actions. 
The medium datasets total approximately 1 million steps across all their trajectories. 
The composition of the medium-replay dataset varies by task: for Hopper, it includes approximately 400,000 steps (comprising states, actions, and rewards) across 2,041 trajectories; for Walker2d, it contains about 300,000 steps within 1,093 trajectories; and for HalfCheetah, approximately 200,000 steps are provided within 202 trajectories.

\subsection{Hyperparameters and Training Setup}
The Decision Transformer model~\cite{chen2021decision} was used for this study. We set the pre-training steps to 5,000, the number of exploration episodes to 1,500, and the training steps per exploration to 300.
We also set the number of training steps for DeFog and ODT so as to be the same as the proposed OnDeFog for fair comparison. The target return and frame drop rates were individually configured for each dataset. Following the experimental settings in DeFog paper~\cite{hudecision}, we set $p_{\mathrm{train}}=0.5$ for the Hopper task's medium dataset and $p_{\mathrm{train}}=0.8$ for other datasets.
Furthermore, the target return was consistent across both the medium and medium-replay datasets, set at 3600 for the Hopper task, 5000 for the Walker2d task, and 6000 for the Halfcheetah task.
Other settings basically follow the ODT paper~\cite{zheng2022online}.

\subsection{Evaluation Method}
For each method, 100 policy evaluations were conducted using models trained with five different random seeds in environments configured with frame drop rates ranging from $0$ to $0.9$ in increments of $0.1$. 
Subsequently, the Interquartile Mean (IQM) of the collected episode returns and their respective 95\% confidence intervals were calculated. 
The adoption of IQM and 95\% confidence intervals for analyzing RL experimental results aligns with recommendations from existing research~\cite{agarwal2021deep}.

\begin{figure}[t]
    \centering
    \includegraphics[width=\linewidth]{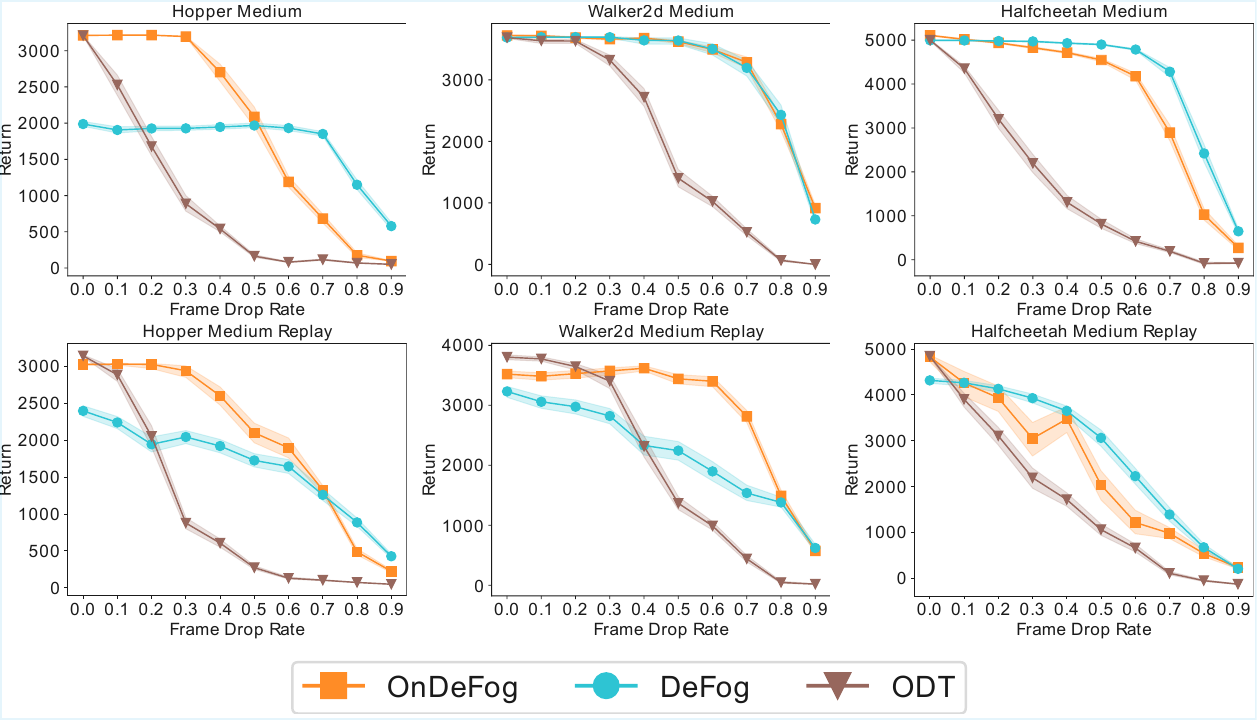}
    \caption{IQMs of return for varying frame drop rates and their 95\% confidence intervals}
    \label{result1}
\end{figure}

\subsection{Comparative Experiments}
This experiment was conducted to perform a comparative analysis of the proposed OnDeFog against the existing DeFog and ODT. The evaluation results for different datasets and frame drop rates are presented in Figure~\ref{result1}.

A comparison between OnDeFog and ODT revealed that ODT exhibits a sharp decline in performance with increasing frame drop rates, while OnDeFog demonstrated a more gradual performance degradation. This observation confirms that OnDeFog is more robust to frame dropping than ODT, likely attributable to its integrated frame dropping handling mechanism. However, OnDeFog demonstrated performance comparable to or slightly inferior to ODT in environments with low frame drop rates. We consider that this outcome is due to the relatively high drop rates employed during OnDeFog's training, which may have led to insufficient adaptation to low drop rate environments. Consequently, this performance disparity is expected to diminish if the actual drop rate in the environment is known a priori.

Compared to DeFog, OnDeFog achieved comparable or superior performance on the medium-replay dataset, as seen in the second row in Figure~\ref{result1}. This performance gap can be attributed to the dominance of low-reward trajectories in the medium-replay dataset, which makes it difficult for DeFog, an offline RL method, to learn high-performing policies. 
OnDeFog achieved relatively high rewards on the medium-replay dataset by acquiring high-reward trajectories through online learning, which facilitated appropriate updates to the replay buffer. 
However, OnDeFog performs inferior to DeFog when applied to the HalfCheetah medium-replay dataset.
This result is likely attributable to HalfCheetah occupying a more expansive action space than other tasks, making exploration insufficient. 
This observation indicates the limitation of OnDeFog's exploration capabilities. 
On the medium dataset, OnDeFog generally performed slightly inferior to DeFog for tasks other than Hopper. 
This outcome is likely due to OnDeFog's online learning exploration failing to discover trajectories with rewards higher than those already present in the medium dataset. As a result, the quality of the replay buffer may have degraded.

These results indicate that, while the proposed OnDeFog can effectively mitigate the performance degradation by leveraging trajectories obtained through online exploration even with datasets containing insufficient reward trajectories, it also encounters challenges such as limited exploration capability when faced with datasets containing extremely low reward trajectories or large action spaces.

\section{Conclusion}
In this study, we proposed OnDeFog, an online RL method that integrates a frame drop learning mechanism, derived from DeFog, into ODT. This integration enables stable learning in frame dropping environments, a significant challenge for conventional ODT. 

In the evaluation experiments, OnDeFog's performance was comprehensively compared with that of existing methods across various frame drop rates. The comparative analysis demonstrated that OnDeFog outperformed the existing ODT method in high-drop-rate environments, thereby confirming its superior robustness to frame dropping. 
Furthermore, when evaluated on the medium-replay datasets, characterized by a high proportion of low-reward data points, OnDeFog often exhibited enhanced performance over DeFog, an offline RL approach. This advantage stemmed from OnDeFog's effective utilization of trajectories obtained via online exploration. However, a challenge emerged during training on the medium dataset, which inherently contains high-reward trajectories: OnDeFog's performance decreased due to its limited ability to discover trajectories with higher rewards.

In future work, a key challenge involves developing a method for automatically adjusting the timing of switching between offline and online learning. By dynamically adjusting this transition based on the quality of the pre-training dataset and the performance of the policy model, we aim to mitigate performance degradation when utilizing low-quality datasets.

\subsubsection*{Acknowledgements}
This work was partially supported by JSPS KAKENHI (JP23H00491 and JP23H03466) and JST PRESTO (JPMJPR2133).

\bibliographystyle{splncs04}
\bibliography{mybib}

\clearpage

\appendix
\section{Freeze-Trunk Fine-Tuning} \label{apdx:sec:ft}
Freeze-trunk fine-tuning, which is a fine-tuning procedure incorporated in DeFog, can be employed in Algorithm~\ref{OnDeFog}. After the pre-defined number of parameter updates, $R-R_{\mathrm{freeze}}$, all parameters are frozen except those of the drop span encoder and the neural network performing action inference. Subsequently, the dropping rate is changed to $p_{\mathrm{ft}}$, and additional training specialized for handling frame dropping is performed.

\section{Details of Model Architecture and Hyperparameters}
We describe the complete list of OnDefog's hyperparameters.
The specific model architecture and its associated hyperparameters are detailed in Table~\ref{hyparamodel}.
Training iterations and dataset-specific hyperparameters are summarized in Tables \ref{hyparanum} and \ref{hyparadata}, respectively. 
During freeze-trunk fine-tuning, model parameters are fixed once the remaining exploration steps reach $R_{\mathrm{freeze}}$, as described in Section~\ref{apdx:sec:ft}. This setting ensures that the total number of training iterations remains consistent, irrespective of whether freeze-trunk fine-tuning is applied.

\begin{table}[tb]
\centering
\caption{Model architecture and hyperparameters of Decision Transformer}
\label{hyparamodel}
\begin{tabular}{ll}
\hline
Hyperparameter          & Value    \\
\hline
Number of Layers        & 4        \\
Number of Attention Heads & 4        \\
Number of Embedding Dimensions     & 512      \\
Training Context Length $K$      & 20       \\
Dropout Rate            & 0.1      \\
Activation Function     & ReLU     \\
Gradient Clipping Norm  & 0.25     \\
Learning Rate           & 0.0001   \\
Weight Decay            & 0.001    \\
Replay Buffer Size $N$      & 1000     \\
Batch Size $B$             & 256      \\
Maximum steps per Episode  & 1000     \\
Optimizer  & \textsc{Lamb}~\cite{LAMB}     \\
\hline
\end{tabular}

\end{table}
\begin{table}[tb]
\centering
\caption{Hyperparameters for model training}
\label{hyparanum}
\begin{tabular}{ll}
\hline
Hyperparameter                          & Value    \\
\hline
Pre-training Steps $I_{\mathrm{offline}}$ & 5000     \\
Number of Exploration Episodes $R$      & 1500     \\
Training Steps per Exploration $I_{\mathrm{online}}$ & 300      \\
\hline
\end{tabular}
\end{table}

\begin{table}[tb]
\centering
\caption{Dataset-specific hyperparameter configurations}
\label{hyparadata}
\begin{tabular}{ccccc}
\hline
Task        & Dataset       & Target Return & Training Drop Rate & FT Drop Rate \\ 
\hline
Hopper      & medium        & 3600          & $p_{\mathrm{train}} = 0.5$                      & $p_{\mathrm{ft}} = 0.8$                      \\
Hopper      & medium-replay & 3600          & $p_{\mathrm{train}} = 0.8$                      & $p_{\mathrm{ft}} = 0.8$                      \\
Walker2d    & medium        & 5000          & $p_{\mathrm{train}} = 0.8$                      & $p_{\mathrm{ft}} = 0.8$                      \\
Walker2d    & medium-replay & 5000          & $p_{\mathrm{train}} = 0.8$                      & $p_{\mathrm{ft}} = 0.8$                      \\
HalfCheetah & medium        & 6000          & $p_{\mathrm{train}} = 0.8$                      & $p_{\mathrm{ft}} = 0.8$                      \\
HalfCheetah & medium-replay & 6000          & $p_{\mathrm{train}} = 0.8$                      & $p_{\mathrm{ft}} = 0.8$                      \\
\hline
\end{tabular}
\end{table}

\section{Supplementary Results} \label{sec:suppl_result}
We compare OnDeFog with its ablated variants, excluding one of the following components: frame dropping, drop-span embedding, or freeze-trunk fine-tuning.
This experiment conducted a comparative analysis of OnDeFog against the ablated components.
Note that we denote ``OnDeFog'' as the method with the freeze-trunk fine-tuning in this section, although it means OnDeFog without the freeze-trunk fine-tuning in the main text.

\begin{figure}[t]
    \centering
    \includegraphics[width=0.8\linewidth]{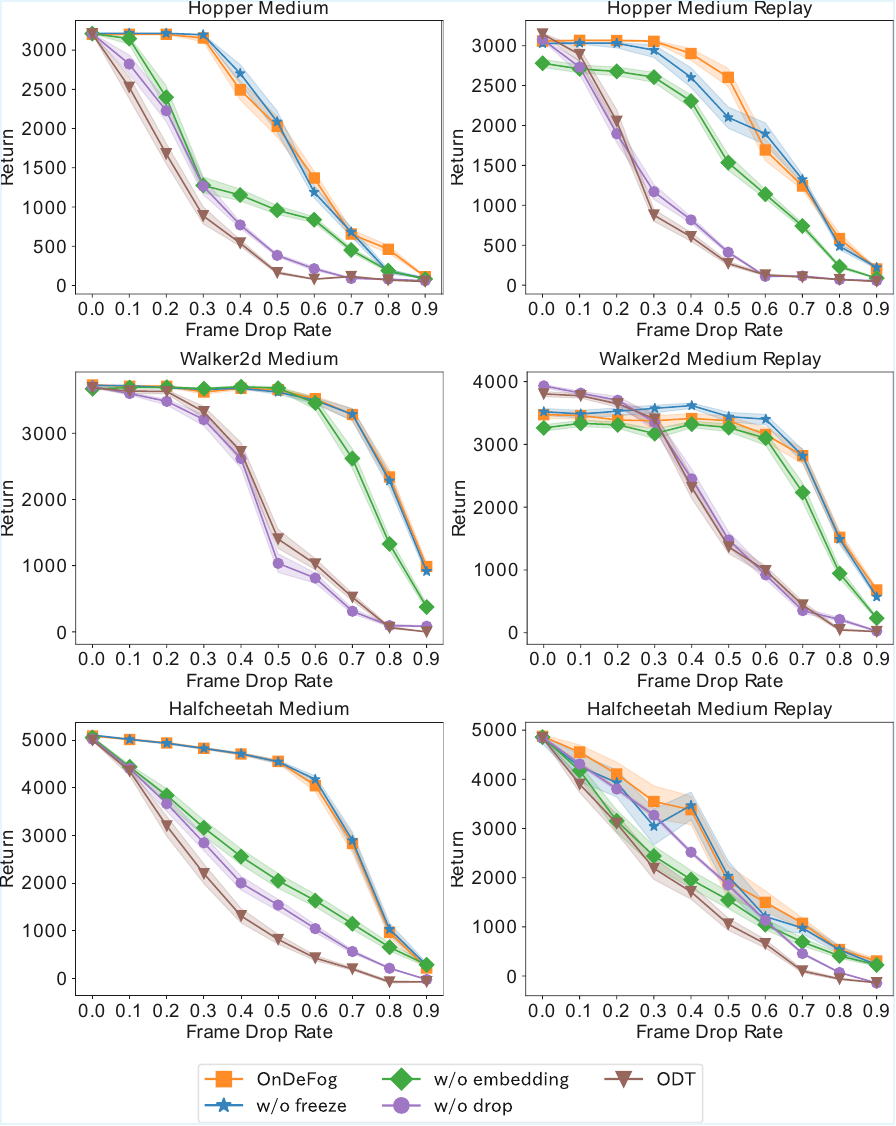}
    \caption{IQMs of return for frame drop rates and their 95\% confidence intervals}
    \label{result2}
\end{figure}

Figure \ref{result2} presents the evaluation results for each dataset across various frame drop rates. In this context, \texttt{w/o drop}, \texttt{w/o embedding}, and \texttt{w/o freeze} denote OnDeFog models without train-time frame dropping, drop-span embedding, and freeze-trunk fine-tuning, respectively. ODT serves as a baseline model lacking all these aforementioned components.

Comparing \texttt{w/o embedding} and OnDeFog, the performance of \texttt{w/o embedding} in the Hopper medium dataset and HalfCheetah medium dataset significantly deteriorated with the drop rate around 0.5, indicating that the stability of training varied across datasets without drop-span embedding. 
Conversely, although \texttt{w/o embedding} exhibited slightly inferior performance compared to OnDeFog, their performance gap is small in the Walker2d task (both datasets) and the Hopper medium-replay dataset. 
This result suggests that, in some possible applications of OnDeFog, the adaptation to frame dropping environments may be achievable even without drop-span embedding, primarily due to train-time frame dropping.

Comparing \texttt{w/o drop} and OnDeFog, \texttt{w/o drop} exhibited comparable or superior performance to OnDeFog in low frame drop rate environments, especcially with the medium-replay datasets. As training for \texttt{w/o drop} was conducted in environments without dropping, its performance at low frame drop rates was similar to that of ODT. However, since drop-span embedding cannot be trained in no-drop environments, \texttt{w/o drop}'s performance declined as the frame drop rate increased, demonstrating an apparent lack of robustness to frame dropping environments.

In the comparison between \texttt{w/o freeze} and OnDeFog, the performance difference was less pronounced than in variants missing other components. The DeFog paper~\cite{hudecision} reported that the freeze-trunk fine-tuning contributes to performance improvement in high-drop-rate environments. 
OnDeFog and \texttt{w/o freeze} exhibit comparable performance, especially evident on the Walker2d medium and HalfCheetah medium datasets. This implies that freeze-trunk fine-tuning has a diminished effect compared to other methods.
More detailed analysis for the effect of freeze-trunk fine-tuning is left as future work.

Based on these findings, train-time frame dropping in the proposed OnDeFog is crucial for adapting to dropping environments. Additionally, drop-span embedding is indispensable for enhancing stable learning in the presence of dropping. Meanwhile, freeze-trunk fine-tuning demonstrated limited effectiveness compared to other components.
\end{document}